\documentclass{article}
\usepackage{iclr2027_conference,times}
\usepackage{microtype}
\usepackage{amsmath,amssymb}
\usepackage{booktabs}
\usepackage{graphicx}
\usepackage{float}
\usepackage{xcolor}
\usepackage{enumitem}
\usepackage{etoolbox}
\usepackage{hyperref}
\usepackage{url}
\usepackage[nameinlink,noabbrev]{cleveref}
\hypersetup{hidelinks}

\AtBeginEnvironment{table}{\setlength{\belowcaptionskip}{6pt}}
\AtBeginEnvironment{table*}{\setlength{\belowcaptionskip}{6pt}}

\floatstyle{ruled}
\newfloat{algorithm}{tbp}{loa}
\floatname{algorithm}{Algorithm}
\crefname{algorithm}{algorithm}{algorithms}

\newcommand{\method}{SCOPE-OPSD}
\newcommand{\opsloss}{\mathcal{L}_{\mathrm{OPSD}}}
\newcommand{\hidloss}{\mathcal{L}_{\mathrm{hid}}}
\newcommand{\totalLoss}{\mathcal{L}_{\mathrm{total}}}
\newcommand{\sg}{\operatorname{sg}}

\iclrfinalcopy

\title{SCOPE-OPSD: Fisher-Conditioned\\
Privileged Subspaces for\\
On-Policy Self-Distillation}
\author{Yunmeng Chen$^{1,*}$ \quad Kunyu Wang$^{2,*}$ \quad Peihan Li$^{1,*}$ \\
Yi Wang$^{1}$ \quad Shuyin Xia$^{3}$ \quad Yi Liu$^{1}$ \quad Xinyong Cheng$^{2}$ \\
Dehui Wang$^{2}$ \quad Xiangyong Zhai$^{2}$ \quad Yanxing Liu$^{1}$ \quad Song Liu$^{1,\dagger}$ \\[0.45em]
\normalfont $^{1}$Chongqing Ant Consumer Finance Co., Ltd. \\
\normalfont $^{2}$Alibaba Cloud Computing Co., Ltd. \\
\normalfont $^{3}$Chongqing University of Posts and Telecommunications \\[0.4em]
\normalfont $^{*}$Equal contribution. \quad $^{\dagger}$Corresponding author. \\[0.3em]
\normalfont\footnotesize
\parbox{\textwidth}{%
\raggedright
\href{mailto:chenyunmeng.cym@myxiaojin.cn}{chenyunmeng.cym@myxiaojin.cn}, \
\href{mailto:wangkunyu.wky@alibaba-inc.com}{wangkunyu.wky@alibaba-inc.com}, \
\href{mailto:oliver.lph@myxiaojin.cn}{oliver.lph@myxiaojin.cn}, \
\href{mailto:haonan.wy@myxiaojin.cn}{haonan.wy@myxiaojin.cn}, \
\href{mailto:xiasy@cqupt.edu.cn}{xiasy@cqupt.edu.cn}, \
\href{mailto:larry.liuy@myxiaojin.cn}{larry.liuy@myxiaojin.cn}, \
\href{mailto:xinyong.cxy@alibaba-inc.com}{xinyong.cxy@alibaba-inc.com}, \
\href{mailto:hanze.wdh@alibaba-inc.com}{hanze.wdh@alibaba-inc.com}, \
\href{mailto:huojing.zxy@alibaba-inc.com}{huojing.zxy@alibaba-inc.com}, \
\href{mailto:liuyanxing.lyx@myxiaojin.cn}{liuyanxing.lyx@myxiaojin.cn}, \
\href{mailto:yushi.ls@myxiaojin.cn}{yushi.ls@myxiaojin.cn}%
}}

\begin{document}
\maketitle
\lhead{Preprint. Work in progress.}

\begin{abstract}
On-policy self-distillation (OPSD) scores student-generated prefixes with a
solution-conditioned self-teacher, yet transfers supervision only through
next-token probabilities. We ask whether the aligned final-layer discrepancy
offers a useful second channel, and how to test that channel without confusing
its geometry with auxiliary strength. \method{} projects the privileged
teacher--student residual onto a frozen rank-64 factor estimated from residual
covariance and language-model-head Fisher sensitivity. It reuses the forwards
already required by OPSD and adds neither rollouts nor inference-time modules.
A matched Random control preserves the structured factor's rank and nonzero
spectrum and uses per-arm gradient-RMS calibration, isolating the effect of
the data-dependent orientation. Across the complete
25/50/75/100-step trajectories for Qwen3-1.7B, 4B, and 8B, Structured is never
below Pure OPSD, with strict gains in 11 of the 12 model--checkpoint
combinations and an exact tie at 4B step 25. Structured also exceeds matched
Random in 10 of the 12 combinations. At step 75 on Qwen3-1.7B, Structured
exceeds matched Random by 1.39 Macro Avg@12 points in each of two independent
training reruns. A cross-fitted diagnostic also
shows 4.40$\times$ greater held-out privileged-gap capture than the matched
random orientation. The results support a compact, Fisher-conditioned
privileged subspace for short-budget OPSD.
\end{abstract}

\section{Introduction}
\label{sec:introduction}

On-policy distillation trains a student on the states it actually visits
\citep{agarwal2024gkd,song2026survey}. This is especially relevant to
long-form reasoning, where a fixed teacher solution may look quite different
from a trajectory sampled by the student. On-Policy Self-Distillation (OPSD)
uses one model under two information conditions \citep{zhao2026opsd}: the
student sees only the problem, whereas a frozen self-teacher also receives a
verified solution. At each student-generated prefix, the privileged teacher
provides a dense next-token target over the full vocabulary. The method is
simple and effective under a short post-training budget.

OPSD, however, uses privileged teacher information only through the token
probabilities produced by the language-model head. The final hidden
states used to produce those probabilities are already aligned by token and
available in the same forward passes, but are otherwise discarded. Recent
work has shown that hidden-space supervision can benefit on-policy
distillation. OPRD aligns representations across layers \citep{yang2026oprd},
and PHF augments OPSD with transition and trajectory geometry
\citep{li2026phf}. A basic attribution question remains: does a hidden
auxiliary help because its orientation captures privileged information, or
because any auxiliary gradient regularizes the student?

We study the narrowest hidden interface already present in OPSD: the final
state immediately before the language-model head. At each rollout position,
we subtract the stopped-gradient privileged-teacher state from the student
state and project the residual through a fixed low-rank factor. The factor is
estimated once from two calibration statistics. Privileged-gap covariance
marks directions in which the two information conditions differ; output
Fisher geometry measures how local hidden perturbations are expressed through
the frozen language-model head. A smooth spectral filter combines them while
suppressing both output-invisible directions and the most strongly expressed
directions already emphasized by the output loss.

\begin{figure*}[t]
    \centering
    \includegraphics[width=\textwidth]{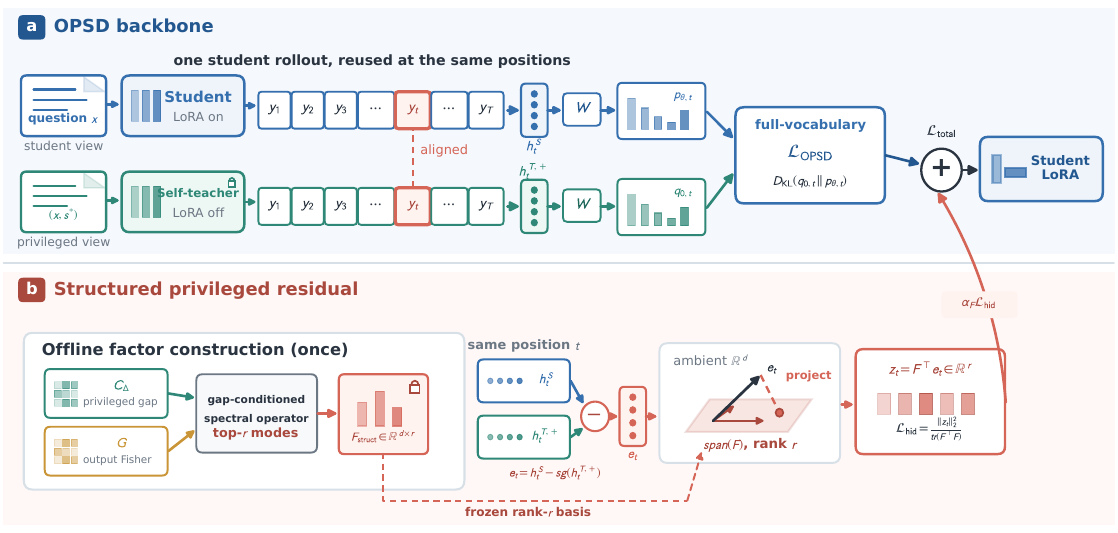}
    \caption{Overview of \method{}. The blue and green paths are the existing
    OPSD computation: one student rollout is rescored by the student and a
    frozen solution-conditioned self-teacher at the same token positions. The
    coral branch projects their final-state residual through a frozen rank-$r$
    factor $F_{\mathrm{struct}}$ and adds the projected loss to full-vocabulary
    OPSD. The factor is fitted once from privileged-gap covariance $C_\Delta$
    and LM-head Fisher geometry $G$. Only the student LoRA is updated.}
    \label{fig:method-overview}
\end{figure*}

The control is as important as the auxiliary. We replace the structured
factor with a random orientation that preserves its rank and nonzero spectrum,
then calibrate each hidden-loss coefficient to the same initial gradient RMS.
Structured versus Random therefore tests the learned orientation while holding
the principal scale variables fixed. Pure OPSD provides the output-only
reference.

We use step 75 as the common checkpoint across model scales. At this checkpoint,
Structured improves matched
Random by 1.39, 0.37, and 0.93 percentage points on Qwen3-1.7B, 4B, and 8B,
respectively; the corresponding gains over Pure OPSD are 1.85, 1.67, and 0.83
points. We also report every saved checkpoint through step 100 rather than
selecting a separate point for each benchmark. Finally, a cross-fitted offline
test shows that the structured factor captures 4.40$\times$ the held-out
privileged-gap score of its matched random control.

Our contributions are:
\begin{itemize}[leftmargin=*,itemsep=2pt]
    \item a final-layer privileged-residual objective that reuses OPSD's
    aligned forwards and adds no rollout or inference-time component;
    \item a frozen Fisher-conditioned rank-64 subspace, paired with spectrum-
    and gradient-matched controls that isolate its orientation; and
    \item matched evidence across three Qwen3 scales, including complete
    checkpoint curves and component controls that
    place GapOnly between Random and Structured while favoring rank 64 over the
    tested ranks 32 and 128 and full-state matching.
\end{itemize}

\section{Background and Problem Formulation}
\label{sec:background}

Let $x$ be a question, $s$ a verified reference solution, and
$y=(y_1,\ldots,y_T)$ a trajectory sampled from the current student
$p_\theta$:
\begin{equation}
    y \sim p_\theta(\cdot\mid x).
\end{equation}
At position $t$, the student predicts from $(x,y_{<t})$. A frozen privileged
self-teacher $q_0$ receives the same student prefix as well as $s$. Following
classical distribution matching and its on-policy extension
\citep{hinton2015distilling,agarwal2024gkd}, OPSD minimizes a token-averaged
forward divergence \citep{zhao2026opsd},
\begin{equation}
\label{eq:opsd}
    \opsloss
    = \mathbb{E}_{(x,s),y}
      \left[
      \frac{1}{T}\sum_{t=1}^{T}
      D_{\mathrm{KL}}\!\left(
      q_0(\cdot\mid x,s,y_{<t})
      \,\Vert\,
      p_\theta(\cdot\mid x,y_{<t})
      \right)
      \right].
\end{equation}
Our implementation follows the original full-vocabulary objective and applies
the same pointwise upper clip of $0.05$ to individual vocabulary
contributions. The teacher remains the step-zero base model with the trainable
student adapter disabled.

Write $h_t^S,h_t^{T,+}\in\mathbb{R}^{d}$ for the final pre-head hidden states of
the student and privileged teacher. With language-model head
$W\in\mathbb{R}^{|\mathcal{V}|\times d}$, OPSD observes these states through
$Wh_t^S$ and $Wh_t^{T,+}$. This output projection is appropriate for next-token
matching, but it leaves two questions unanswered. First, does the privileged
residual
\begin{equation}
    e_t = h_t^S - \sg(h_t^{T,+})
\end{equation}
carry training information not already exploited by \cref{eq:opsd}? Second,
if it does, which directions of $e_t$ should be supervised?

The second question creates a confound. Hidden losses with identical scalar
coefficients can have very different gradient magnitudes because their
operators have different spectra and align differently with the model
Jacobian. A fair orientation comparison therefore requires more than assigning
the same coefficient to two projectors. We explicitly calibrate the initial
auxiliary gradient and treat orientation quality as a separate hypothesis from
the usefulness of hidden supervision itself.

\section{Fisher-Conditioned Privileged Supervision}
\label{sec:method}

\subsection{A minimal final-layer auxiliary}

\Cref{fig:method-overview} summarizes the separation between the existing
OPSD computation, the one-time factor construction, and the online objective.

For a fixed factor $F\in\mathbb{R}^{d\times r}$, we define
\begin{equation}
\label{eq:hidden-loss}
    \hidloss(F)
    = \mathbb{E}_{(x,s),y}
      \left[
      \frac{1}{T}\sum_{t=1}^{T}
      \frac{\lVert F^\top e_t\rVert_2^2}
           {\operatorname{tr}(F^\top F)}
      \right].
\end{equation}
The teacher state is stop-gradient. The trace normalization makes the reported
loss less sensitive to a uniform rescaling of $F$; it does not by itself match
the gradient delivered to the trainable adapter. The training objective is
\begin{equation}
\label{eq:total-loss}
    \totalLoss = \opsloss + \alpha_F\hidloss(F),
\end{equation}
where $\alpha_F$ is calibrated for each control as described below.

The auxiliary uses the same token positions and forwards as OPSD. Its only new
online operation is the matrix multiplication $F^\top e_t$; it requires no
additional model forward, rollout, reward model, or inference-time state.

\subsection{A gap- and sensitivity-conditioned factor}

We construct the structured factor offline from a calibration set. Let
$q_t=q_0(\cdot\mid x,s,y_{<t})$ and let $W$ be frozen. In the
information-geometric view of local model sensitivity \citep{amari1998natural},
the output Fisher pulled back to hidden space is
\begin{equation}
\label{eq:fisher}
    G_t = W^\top
    \left[\operatorname{Diag}(q_t)-q_tq_t^\top\right]W,
    \qquad
    G=\mathbb{E}_t[G_t].
\end{equation}
At each position, the implementation estimates $G_t$ from 16 sampled LM-head
rows rather than forming the full-vocabulary covariance. If
$G=U\operatorname{Diag}(\lambda)U^\top$, we apply the smooth spectral weight
\begin{equation}
\label{eq:bandpass}
    a_\mu(\lambda)
    = \frac{\lambda}{\mu}
      \exp\!\left(1-\frac{\lambda}{\mu}\right),
\end{equation}
where $\mu$ is the geometric median of the positive eigenvalues. This weight
suppresses directions that are nearly invisible to the output head and
directions already carrying the largest local output sensitivity.

We also estimate the privileged residual covariance
\begin{equation}
    C_\Delta = \mathbb{E}_t[e_te_t^\top].
\end{equation}
In the Fisher eigenbasis, define
\begin{equation}
    \widetilde{M}
    = \operatorname{Diag}(\sqrt{a})
      U^\top C_\Delta U
      \operatorname{Diag}(\sqrt{a}).
\end{equation}
If $V_r$ contains the top $r$ eigenvectors of $\widetilde{M}$, the structured
factor is
\begin{equation}
\label{eq:factor}
    F_{\mathrm{struct}}
    = U\operatorname{Diag}(\sqrt{a})V_r.
\end{equation}
We use $r=64$. The factor is fitted once and remains frozen during training.
We call this operator Structured Covariance and Output-Fisher Projected Error
(SCOPE), and the resulting OPSD method \method{}.

\begin{algorithm}[H]
\caption{\method{} calibration and training}
\label{alg:scope}
\textbf{Input:} OPSD data $\mathcal{D}=\{(x,s)\}$, frozen base model $q_0$,
rank $r$, and target hidden-gradient RMS $\gamma$.
\begin{enumerate}[leftmargin=1.45em,itemsep=1pt,topsep=3pt]
    \item \textbf{Calibrate once.} On 128 calibration prompts, sample aligned
    positions from student trajectories; record $e_t$, $q_t$, and LM-head rows.
    \item Estimate $C_\Delta$ and $G$; apply \cref{eq:bandpass}; compute the
    top-$r$ factor $F_{\mathrm{struct}}$ using \cref{eq:factor}.
    \item Construct $F_{\mathrm{rand}}$ with the same nonzero spectrum but an
    independent orthonormal orientation.
    \item For each factor $F$, measure
    $\operatorname{RMS}(\nabla_\theta\hidloss(F))$ at initialization and set
    $\alpha_F$ to match $\gamma$.
    \item \textbf{Train.} For each optimizer step, sample one student rollout,
    evaluate student and frozen privileged teacher on its aligned prefixes,
    compute \cref{eq:opsd,eq:hidden-loss}, and update only the student LoRA with
    \cref{eq:total-loss}.
\end{enumerate}
\textbf{Output:} one deployable student checkpoint; discard $F$ at inference.
\end{algorithm}

\subsection{Matched orientation control}

Our primary control replaces the orientation of
$F_{\mathrm{struct}}$ while preserving its nonzero spectrum. In particular,
write
\begin{equation}
    F_{\mathrm{struct}}F_{\mathrm{struct}}^\top
    = P\operatorname{Diag}(\sigma_1^2,\ldots,\sigma_r^2)P^\top.
\end{equation}
For an orthonormal frame $Q\in\mathbb{R}^{d\times r}$ satisfying
$Q^\top Q=I_r$, the matched random factor is
\begin{equation}
\label{eq:matched-random}
    F_{\mathrm{rand}}=Q\operatorname{Diag}(\sigma_1,\ldots,\sigma_r).
\end{equation}
It therefore has the same rank, trace, and nonzero spectrum as the structured
factor, but its orientation is independent of the privileged residual.
We sample $Q$ independently of the privileged data from an orthonormal frame;
this control construction supports arbitrary hidden dimension $d$.

Equal trace does not imply equal parameter updates because the two factors may
align differently with the model Jacobian. At initialization, let
\begin{equation}
    g_F = \operatorname{RMS}\!\left(
        \nabla_\theta \hidloss(F)
    \right).
\end{equation}
For a common target $\gamma$, we set
\begin{equation}
\label{eq:gradient-calibration}
    \alpha_F = \gamma/g_F.
\end{equation}
Pure OPSD sets $\alpha_F=0$. Structured and Random are alternative
orientations of the same rank-$r$ auxiliary, rather than additive components;
their matched comparison is the main test of the proposed geometry.

\subsection{Local interpretation of the structured factor}

The construction in \cref{eq:fisher,eq:factor} ranks directions using a local
proxy: privileged discrepancy weighted by sensitivity through the output head.
The offline diagnostic tests whether the fitted factor captures this target on
held-out prompts. Downstream evaluation separately tests whether the local
orientation improves autoregressive reasoning.

\section{Experimental Protocol}
\label{sec:experiments}

\paragraph{Compute environment.}
Every training run uses one 16-device Alibaba T-Head Zhenwu 810E (ZW810E)
PPU node, with 96~GB HBM2e per device. The actor is sharded across all devices
with FSDP. We use PyTorch 2.9.0 and the project VERL snapshot with its
PPU-compatible backend; evaluation is served with SGLang. All matched arms at a
given model scale use the same hardware class and software environment.

\paragraph{Training data and OPSD recipe.}
We use the 29,434-example OpenThoughts Math OPSD split
\citep{guha2025openthoughts,zhao2026opsddataset} and Qwen3-1.7B, 4B, and 8B
\citep{yang2025qwen3}. The student samples one non-thinking response per
problem with temperature 1.1, top-p 0.95, top-k 20, and a 1,024-token limit.
The privileged teacher is the frozen step-zero model in thinking mode; it
receives the verified solution and the same student prefix. We follow the OPSD
recipe: full-vocabulary forward KL, pointwise clip 0.05, LoRA rank 64 and alpha
128 \citep{hu2022lora}, learning rate $5\times10^{-6}$, and effective batch
size 32. Each arm runs for 100 optimizer steps and saves steps 25, 50, 75, and
100.

\paragraph{Scale-specific calibration.}
At each scale, we estimate $G$, $C_\Delta$, and a rank-64 factor from 128
calibration prompts. Factors are neither padded nor transferred between
models. The hidden widths are 2,048, 2,560, and 4,096. We also recalibrate the
auxiliary coefficient at each scale using \cref{eq:gradient-calibration}; the
rank, training schedule, and evaluation protocol remain unchanged.

\paragraph{Compared systems.}
Base is the untrained model. Pure OPSD uses only \cref{eq:opsd}. Matched Random
uses \cref{eq:matched-random}, and Structured uses
\cref{eq:factor}. Within a matched run, Pure, Random, and Structured use the
same training split, batch and checkpoint schedules, rollout configuration,
frozen teacher, adapter, optimizer, and output objective. Random and Structured
receive the same initial hidden-gradient RMS. The comparison therefore adds no
rollout samples, optimizer steps, or privileged inputs beyond Pure OPSD.

We additionally train a local Group Relative Policy Optimization (GRPO)
reference under the source-paper recipe for 500 steps, saving checkpoints
100, 200, 300, 400, and 500 \citep{zhao2026opsd}. GRPO changes both the
objective and the schedule, so it is a standard post-training context row, not
a control for the Pure--Random--Structured comparison.

\paragraph{Ablation variants.}
At Qwen3-1.7B and fixed step 75, GapOnly fits a rank-64 factor from privileged
residual covariance without Fisher conditioning. FullHidden penalizes all
$d=2{,}048$ final-state coordinates. The rank sweep applies the complete
Structured construction with $r\in\{32,64,128\}$. These variants retain the
same OPSD output objective and evaluation protocol. They test component
ordering and selective capacity; they are not used to retune the frozen
cross-scale comparison.

\paragraph{Checkpoint policy.}
Based on the Qwen3-1.7B checkpoint trajectory, we use step 75 as the common
checkpoint for the primary cross-scale comparison and apply the same choice to
Qwen3-4B and Qwen3-8B. We additionally report all saved checkpoints at steps 25,
50, 75, and 100 to characterize convergence. Within each comparison, every
benchmark uses the same checkpoint; we do not construct benchmark-wise
checkpoint oracles.

This distinction matters for external context. The original OPSD paper reports
benchmark-wise best values drawn from its checkpoint sweep, which need not
correspond to one deployable model; it separately provides a single step-100
value for Qwen3-1.7B \citep{zhao2026opsd}. PHF reports an independent
step-100 OPSD reproduction \citep{li2026phf}. We label these selection rules
when discussing external numbers and compute all method deltas only between
our locally matched methods.

\paragraph{Evaluation.}
We follow the OPSD thinking-mode protocol on AIME 2024, AIME 2025, and HMMT
2025, with 30 problems per benchmark. Each problem receives 12 samples at
temperature 1.0 and top-p 0.95, with at most 38,912 new tokens. If
$c_{m,d,i,j}\in\{0,1\}$ denotes correctness for method $m$, benchmark $d$,
question $i$, and sample $j$, then
\begin{equation}
\label{eq:avg12}
    \widehat{A}_{m,d}
    = \frac{100}{30\cdot12}
      \sum_{i=1}^{30}\sum_{j=1}^{12}c_{m,d,i,j},
    \qquad
    \widehat{M}_m=\frac{1}{3}\sum_d\widehat{A}_{m,d}.
\end{equation}
Thus Avg@12 is generation-level accuracy, and the reported Macro is the
unweighted mean across the three benchmarks.

\paragraph{Evaluation integrity.}
Every system is evaluated under the same sampling configuration. Each reported
checkpoint covers all 90 problems with 12 samples per problem, yielding 1,080
generations. Only complete evaluations with zero failed requests enter the
tables. Before aggregation, we verify the actor checkpoint, LoRA merge
provenance, served-model identity, dataset hashes, decoding configuration, and
generation count.

\section{Results}
\label{sec:results}

\subsection{The fixed step-75 comparison transfers across scales}

\begin{table*}[t]
\centering
\small
\setlength{\tabcolsep}{4pt}
\caption{Primary matched comparison at the common step-75 checkpoint under the
OPSD Think Avg@12 protocol. All methods use the same training and evaluation
recipe within each scale; deltas are percentage points.}
\label{tab:cross-scale-s75}
\begin{tabular}{lrrrrrr}
\toprule
Scale & Base & Pure OPSD & \shortstack{Matched\\Random} & Structured
      & Struct.--Pure & Struct.--Random \\
\midrule
Qwen3-1.7B & 37.41 & 41.48 & 41.94 & \textbf{43.33} & +1.85 & +1.39 \\
Qwen3-4B   & 60.65 & 62.13 & 63.43 & \textbf{63.80} & +1.67 & +0.37 \\
Qwen3-8B   & 62.69 & 64.45 & 64.35 & \textbf{65.28} & +0.83 & +0.93 \\
\bottomrule
\end{tabular}
\end{table*}

Structured is the strongest local arm at all three scales in
\cref{tab:cross-scale-s75}. Its advantage over Pure OPSD ranges from 0.83 to
1.85 points, and it also exceeds the spectrum- and gradient-matched random
orientation at every scale. The checkpoint is identical across the table and
is shared by every benchmark and model scale. Even when Pure OPSD is
allowed its best saved checkpoint in \{25,50,75,100\}, fixed step-75 Structured
remains higher by 1.85, 1.21, and 0.65 points at 1.7B, 4B, and 8B.

\subsection{Structured orientation improves over matched random at 1.7B}

\begin{table}[t]
\centering
\caption{Qwen3-1.7B Avg@12 at the common step-75 checkpoint.}
\label{tab:main-results}
\begin{tabular}{lrrr}
\toprule
Dataset & Matched random & Structured & Structured--Random \\
\midrule
AIME24 & 57.78 & \textbf{58.33} & +0.56 \\
AIME25 & 40.70 & \textbf{42.50} & +1.81 \\
HMMT25 & 27.36 & \textbf{29.17} & +1.81 \\
\midrule
Macro & 41.94 & \textbf{43.33} & +1.39 \\
\bottomrule
\end{tabular}
\end{table}

At step 75, Structured improves over Matched Random by 0.56, 1.81, and 1.81
points on AIME24, AIME25, and HMMT25, respectively, yielding a +1.39-point
Macro gain. The improvement is therefore not driven by a single benchmark.
Across the complete checkpoint trajectory, Structured also maintains a higher
Macro score at steps 50, 75, and 100 (\cref{tab:curve-1p7b}).

For context, the original OPSD study reports 43.40 for Qwen3-1.7B after taking
the best checkpoint separately for each benchmark, and 42.50 for its single
step-100 checkpoint \citep{zhao2026opsd}. Our Structured model reaches 43.33
at the common step-75 checkpoint. At 4B and 8B, the corresponding Structured
scores are 63.80 and 65.28, compared with the source paper's benchmark-wise
checkpoint oracles of 63.60 and 64.80. Because the checkpoint-selection rules
differ, these external values provide scale context only and are not used to
compute the matched deltas in \cref{tab:cross-scale-s75}.

\subsection{A local GRPO curve anchors the post-training context}

The locally trained Qwen3-1.7B GRPO curve obtains Macro Avg@12 values of 36.94,
38.24, 39.44, 38.80, and 38.61 at steps 100--500. Its best single deployable
checkpoint is step 300 at 39.44, above the local Base score of 37.41 and below
Pure OPSD at 41.48 and Structured at 43.33. Structured is 3.89 points higher
than this local GRPO checkpoint. This is a contextual comparison rather than a
matched causal delta: GRPO uses a different objective and a 500-step schedule,
whereas the OPSD arms share a 100-step distillation recipe. The full curve and
source-paper context appear in \cref{tab:grpo-curve}.

\subsection{Component and rank controls support selective structure}

\begin{figure*}[t]
\centering
\includegraphics[width=\textwidth]{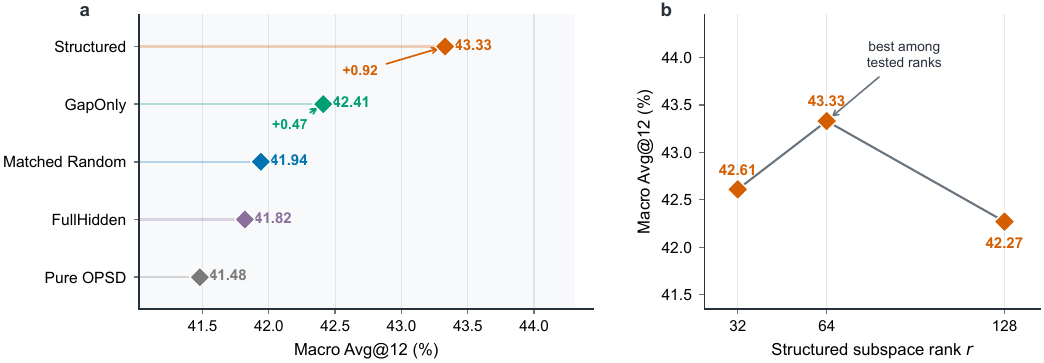}
\caption{Qwen3-1.7B ablations at step 75 under the same Think Avg@12 protocol.
\textbf{(a)} Component ordering and the FullHidden control. \textbf{(b)} Rank
sensitivity of the complete Structured construction. Each marker denotes the
reported method-level aggregate.}
\label{fig:component-rank}
\end{figure*}

\Cref{fig:component-rank} separates two questions. First, the component sequence
is Random $<$ GapOnly $<$ Structured: 41.94 $<$ 42.41 $<$ 43.33. Matched
Random is 0.46 points above Pure OPSD, GapOnly is 0.47 points above Random, and
Structured is another 0.92 points above GapOnly. Under the tested setup, this
ordering supports incremental contributions from a hidden auxiliary,
privileged residual information, and Fisher conditioning.
FullHidden averages 41.82, 0.12 points below Random and 1.51 below Structured;
constraining more coordinates is therefore not monotonically beneficial in
this setting.

Second, rank 64 is the strongest of the tested Structured ranks. It exceeds
ranks 32 and 128 by 0.72 and 1.06 points. Together, these controls support the
use of a privileged, Fisher-conditioned, selective low-rank subspace under the
tested setup; they do not imply a unique causal decomposition or that rank 64
is universally optimal.

\begin{figure*}[t]
\centering
\includegraphics[width=\textwidth]{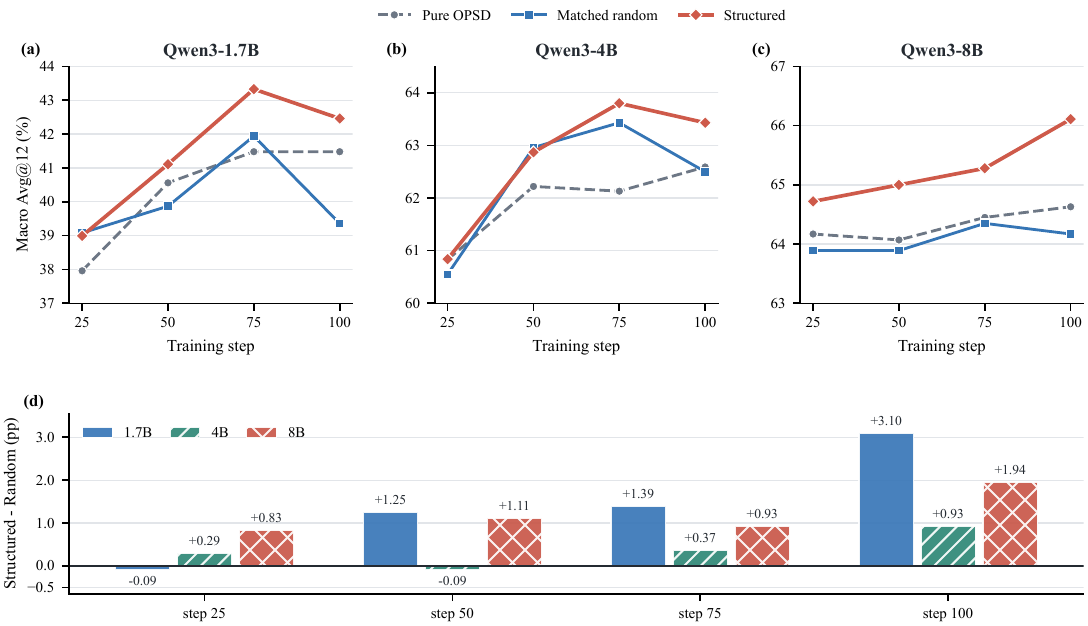}
\caption{Complete checkpoint trajectories. \textbf{(a--c)} Macro Avg@12 at
steps 25, 50, 75, and 100 for Pure OPSD, spectrum- and gradient-matched Random,
and Structured. \textbf{(d)} Structured minus Random for each matched
scale--checkpoint pair. All points use the same method-level reporting
protocol.}
\label{fig:cross-scale-results}
\end{figure*}

\subsection{Complete trajectories and the terminal checkpoint}

Structured is never below Pure OPSD in any of the 12 scale--checkpoint
comparisons in \cref{fig:cross-scale-results}: it is strictly above in 11 and
exactly tied at 4B step 25 (657 versus 657 correct generations). It exceeds
Random in 10 of 12; the exceptions are 1.7B step 25 and 4B step 50, each by
0.09 points. At the terminal step 100,
the Structured--Random differences are +3.10, +0.93, and +1.94 points for
1.7B, 4B, and 8B. The corresponding Structured scores are 42.46, 63.43, and
66.11. Thus the primary step-75 result is not produced by choosing a different
checkpoint for each benchmark, and the ordering persists at the shared budget
endpoint.

The trajectories also show why the terminal checkpoint should not be treated
as universally optimal. Structured peaks at step 75 for 1.7B and 4B, but at
step 100 for 8B. Reporting the full curve separates this convergence behavior
from the common cross-scale comparison.

\subsection{The fitted factor captures its offline target}

On held-out calibration halves, the structured factor captures
4.40$\times$ the privileged-gap score of a spectrum-matched random
orientation. Under a local full-vocabulary intervention, its mean KL reduction
is 0.2347, compared with 0.0241 for Random. These diagnostics verify that the
factor is data dependent and selects the intended local geometry; they do not
replace the downstream test, which measures correctness after autoregressive
rollout. \Cref{app:gate} gives the estimands and complete diagnostic table.

\section{Related Work}
\label{sec:related}

\paragraph{Knowledge and representation distillation.}
Classical distillation matches teacher and student distributions
\citep{hinton2015distilling}. Representation variants transfer intermediate
features, attention, relations, or contrastive structure
\citep{romero2015fitnets,zagoruyko2017attention,tian2020crd}, including hidden
layers in Transformers \citep{sun2019patient,wang2020minilm}. Unlike these
fixed-distribution settings, we compare ordinary and solution-conditioned
views of one model along its own trajectories.

\paragraph{On-policy language-model distillation.}
Generalized Knowledge Distillation trains on student sequences
\citep{agarwal2024gkd}; later work adapts divergences or token weights
\citep{ko2025distillm2,zhang2025aligndistil,jung2025todi,song2026survey}.
OPSD instead gives a frozen self-teacher a verified solution and retains dense
full-vocabulary supervision on student prefixes \citep{zhao2026opsd}.
Subsequent studies examine teacher bias, token dynamics, and supervision scope
\citep{li2026rethinking,jiang2026rock,harne2026biased,zhao2026psopsd,
hou2026dash,yang2026usd}. OP2SD changes the privileged example
\citep{ichihara2026op2sd}, PAST conditions on the completed student trajectory
and outcome \citep{feng2026past}, and SA-OPD filters weakly grounded signals
\citep{jiang2026saopd}. Recent SCOPE calibrates OPD by routing correct and
incorrect rollouts through separate, perplexity-weighted supervision paths
\citep{zheng2026scope}. By contrast, \method{} leaves rollout routing and the
full-vocabulary OPSD objective unchanged, adding an auxiliary loss on a frozen
Fisher-conditioned privileged-residual subspace.

\paragraph{Hidden-space and geometric supervision.}
OPRD aligns selected hidden representations and can replace output-space
distillation \citep{yang2026oprd}. The contemporaneous PHF preprint augments
OPSD with hidden transitions, within-layer trajectory geometry, and
neighboring-layer relations \citep{li2026phf}. Its ablations remove additive
loss components; we instead test one final-layer privileged-residual
orientation against a rank-, spectrum-, and gradient-matched replacement.
Latent On-Policy Self-Distillation (LOPD) makes the teacher's privileged
context learnable end to end as continuous latent tokens, while retaining
dense token-level supervision at each visited prefix
\citep{zhang2026lopd}.
FP-OPD uses Fisher/tangent geometry for an output correction in vision-language
distillation \citep{xue2026fpopd}; we use output Fisher only to construct a
frozen hidden factor for language-model self-distillation. Related analyses
motivate separating local update geometry from sequence-level correctness
\citep{shen2026geometry,ge2026scaling}.

\section{Limitations}
\label{sec:limitations}

Our experiments cover one model family and three competition-mathematics
benchmarks. Other architectures, reasoning domains, and longer training
budgets may exhibit different convergence behavior. Avg@12 measures accuracy
under a fixed sampling budget rather than large-$K$ solvability.

The matched-random comparison controls rank, spectrum, and initial
hidden-gradient scale. The GapOnly ordering separates residual covariance from
Fisher conditioning, while FullHidden and the rank sweep support a selective
low-rank design. These mechanism controls are currently limited to 1.7B. Their
ordering supports the roles of privileged residual information, Fisher
conditioning, and selective rank under the tested setup, but does not establish
a unique causal decomposition or a rank that is optimal beyond the tested set.

\section{Conclusion}
\label{sec:conclusion}

\method{} adds a frozen Fisher-conditioned privileged subspace to the existing
OPSD computation. It reuses aligned student and teacher states, requires no
extra rollout, and leaves deployment unchanged. At the common step-75
checkpoint, Structured improves both Pure OPSD and a spectrum- and
gradient-matched random orientation on Qwen3-1.7B, 4B, and 8B. The complete
checkpoint curves preserve the broader ordering through the shared 100-step
budget and avoid benchmark-specific checkpoint selection. The
cross-fitted diagnostic connects this downstream behavior to the intended
privileged-gap geometry. Component controls place GapOnly between Random and
Structured, while full-state matching and the tested ranks 32 and 128 remain
below rank 64. A locally trained GRPO curve provides a standard post-training
reference without being treated as a matched objective or budget comparison.
Together, these results support a compact hidden channel for short-budget
on-policy self-distillation.

\bibliography{references}
\bibliographystyle{iclr2027_conference}

\clearpage
\appendix
\section{Reproducibility Details}
\label{app:reproducibility}

\subsection{Training configuration}

\begin{table}[H]
\centering
\small
\caption{Shared training configuration.}
\label{tab:train-config}
\begin{tabular}{@{}p{0.42\linewidth}p{0.52\linewidth}@{}}
\toprule
Setting & Value \\
\midrule
Compute node & 16$\times$ Alibaba T-Head Zhenwu 810E (ZW810E) PPUs \\
Device memory & 96~GB HBM2e per PPU \\
Training stack & PyTorch 2.9.0; VERL with PPU-compatible backend; FSDP \\
Evaluation serving & SGLang \\
Training split & OpenThoughts Math OPSD, 29,434 examples \\
Optimizer / saved steps & 100 / \{25, 50, 75, 100\} \\
Effective batch size & 32 \\
Student rollout & non-thinking, one response, at most 1,024 tokens \\
Teacher & thinking, frozen step-zero self-teacher \\
Rollout sampling & temperature 1.1, top-p 0.95, top-k 20 \\
Output objective & full-vocabulary forward KL; clip 0.05 \\
Adapter & LoRA rank 64, alpha 128 \\
Learning rate & $5\times10^{-6}$ \\
Hidden rank / calibration prompts & 64 / 128 \\
\bottomrule
\end{tabular}
\end{table}

\begin{table}[H]
\centering
\small
\caption{Scale-specific factor calibration.}
\label{tab:scale-config}
\begin{tabular}{lrrr}
\toprule
Scale & Hidden / layers & $\alpha_{\mathrm{rand}}$ & $\alpha_{\mathrm{struct}}$ \\
\midrule
1.7B & 2,048 / 28 & $6.3589{\times}10^{-4}$ & $2.5626{\times}10^{-4}$ \\
4B   & 2,560 / 36 & $6.7885{\times}10^{-4}$ & $2.4665{\times}10^{-4}$ \\
8B   & 4,096 / 36 & $8.3017{\times}10^{-4}$ & $3.6154{\times}10^{-4}$ \\
\bottomrule
\end{tabular}
\end{table}

All three models use vocabulary size 151,936, auxiliary rank 64, and a
100-step training budget.

The matched random control samples an orthonormal frame independently of the
privileged data. The implementation uses randomized normalized Hadamard columns
when $d$ is a power of two and reduced QR of a pseudorandom Gaussian matrix
otherwise. Both constructions satisfy $Q^\top Q=I_r$ and affect only the
counterfactual control in \cref{eq:matched-random}; the structured method is
unchanged across hidden widths.

\subsection{Evaluation and statistical estimands}

\begin{table}[H]
\centering
\small
\caption{Evaluation configuration.}
\label{tab:eval-config}
\begin{tabular}{@{}ll@{}}
\toprule
Setting & Value \\
\midrule
Benchmarks & AIME24, AIME25, HMMT25 \\
Problems / samples per problem & 30 / 12 \\
Maximum new tokens & 38,912 \\
Sampling & temperature 1.0, top-p 0.95 \\
Prompt mode & Qwen3 thinking mode \\
Metric & Avg@12; three-benchmark macro \\
Generations per evaluation & 1,080 \\
\bottomrule
\end{tabular}
\end{table}

Equation~\ref{eq:avg12} defines the point estimators. For a matched pair of
methods $a$ and $b$, we report
$\widehat{\Delta}_{a,b}=\widehat{M}_a-\widehat{M}_b$. Tables and figures report
method-level aggregates under the common evaluation protocol.

Every result row has 1,080/1,080 generations and zero failed requests. We also
verify source actor step, LoRA merge provenance, served-model identity, dataset
hashes, and decoding configuration before result aggregation.

\subsection{Analytical incremental cost}

\begin{table}[H]
\centering
\small
\caption{Incremental online cost relative to Pure OPSD. $T$ is response length,
$d$ hidden width, and $r=64$.}
\label{tab:complexity}
\begin{tabular}{@{}ll@{}}
\toprule
Quantity & Increment \\
\midrule
Student rollouts / teacher forwards & 0 / 0 \\
Projection arithmetic & $\Theta(Tdr)$ \\
Projected activations / frozen factor & $O(Tr)$ / $dr$ scalars \\
Trainable parameters & 0 \\
Inference parameters / FLOPs & 0 / 0 \\
\bottomrule
\end{tabular}
\end{table}

The factor contains 131,072, 163,840, and 262,144 scalars at 1.7B, 4B, and 8B.
Offline calibration uses 128 prompts, stores dense $d\times d$ moment matrices,
and performs one dense spectral solve per factor fit. This one-time stage is not
part of online training and the factor is discarded after training. The table
reports analytical properties of the added interface rather than end-to-end
system measurements.

\subsection{Data provenance}

We use the released OPSD training split without adding AIME24, AIME25, or HMMT25
examples. Evaluation files are immutable 30-problem sets with recorded hashes,
preserving the source recipe's train/evaluation boundary.

\section{Complete Checkpoint Curves}
\label{app:curves}

\subsection{Qwen3-1.7B}

\begin{table}[H]
\centering
\small
\caption{Macro Avg@12 across the complete Qwen3-1.7B checkpoint trajectory.}
\label{tab:curve-1p7b}
\begin{tabular}{rrrrrr}
\toprule
Step & Pure & Random & Structured & Struct.--Pure & Struct.--Random \\
\midrule
25  & 37.96 & 39.08 & 38.99 & +1.03 & -0.09 \\
50  & 40.56 & 39.87 & 41.11 & +0.55 & +1.25 \\
75  & 41.48 & 41.94 & \textbf{43.33} & +1.85 & +1.39 \\
100 & 41.48 & 39.36 & 42.46 & +0.98 & +3.10 \\
\bottomrule
\end{tabular}
\end{table}

\subsection{Qwen3-4B}

\begin{table}[H]
\centering
\small
\setlength{\tabcolsep}{3pt}
\caption{Complete Qwen3-4B Avg@12 trajectory.}
\label{tab:curve-4b}
\begin{tabular}{lrrrrr}
\toprule
Arm & Step & AIME24 & AIME25 & HMMT25 & Macro \\
\midrule
Base & 0 & 73.61 & 65.56 & 42.78 & 60.65 \\
\midrule
Pure & 25 & 73.61 & 66.11 & 42.78 & 60.83 \\
Pure & 50 & 74.72 & 67.50 & 44.44 & 62.22 \\
Pure & 75 & 75.56 & 67.78 & 43.06 & 62.13 \\
Pure & 100 & 76.11 & 67.22 & 44.44 & 62.59 \\
\midrule
Random & 25 & 74.44 & 66.39 & 40.83 & 60.55 \\
Random & 50 & 76.67 & 68.33 & 43.89 & 62.96 \\
Random & 75 & 76.11 & 67.78 & 46.39 & 63.43 \\
Random & 100 & 75.56 & 68.06 & 43.89 & 62.50 \\
\midrule
Structured & 25 & 74.17 & 67.78 & 40.56 & 60.84 \\
Structured & 50 & 74.44 & 70.83 & 43.33 & 62.87 \\
Structured & 75 & 75.28 & 69.44 & 46.67 & \textbf{63.80} \\
Structured & 100 & 73.06 & 70.83 & 46.39 & 63.43 \\
\bottomrule
\end{tabular}
\end{table}

\subsection{Qwen3-8B}

\begin{table}[H]
\centering
\small
\setlength{\tabcolsep}{3pt}
\caption{Complete Qwen3-8B Avg@12 trajectory.}
\label{tab:curve-8b}
\begin{tabular}{lrrrrr}
\toprule
Arm & Step & AIME24 & AIME25 & HMMT25 & Macro \\
\midrule
Base & 0 & 76.67 & 65.56 & 45.83 & 62.69 \\
\midrule
Pure & 25 & 76.67 & 71.11 & 44.72 & 64.17 \\
Pure & 50 & 75.83 & 70.83 & 45.56 & 64.07 \\
Pure & 75 & 75.28 & 72.50 & 45.56 & 64.45 \\
Pure & 100 & 77.22 & 70.28 & 46.39 & 64.63 \\
\midrule
Random & 25 & 77.22 & 68.33 & 46.11 & 63.89 \\
Random & 50 & 78.06 & 68.33 & 45.28 & 63.89 \\
Random & 75 & 77.50 & 69.72 & 45.83 & 64.35 \\
Random & 100 & 77.50 & 69.17 & 45.83 & 64.17 \\
\midrule
Structured & 25 & 76.11 & 72.78 & 45.28 & 64.72 \\
Structured & 50 & 78.06 & 72.78 & 44.17 & 65.00 \\
Structured & 75 & 78.61 & 71.39 & 45.83 & 65.28 \\
Structured & 100 & 76.94 & 73.89 & 47.50 & \textbf{66.11} \\
\bottomrule
\end{tabular}
\end{table}

\section{Local GRPO Post-Training Context}
\label{app:grpo}

\begin{table}[H]
\centering
\small
\setlength{\tabcolsep}{4pt}
\caption{Complete local Qwen3-1.7B GRPO checkpoint curve under the official
Think Avg@12 protocol.}
\label{tab:grpo-curve}
\begin{tabular}{rrrrr}
\toprule
Step & AIME24 & AIME25 & HMMT25 & Macro \\
\midrule
100 & 51.11 & 36.94 & 22.78 & 36.94 \\
200 & 52.50 & 38.61 & 23.61 & 38.24 \\
300 & 56.11 & 38.33 & 23.89 & \textbf{39.44} \\
400 & 52.78 & 38.06 & 25.56 & 38.80 \\
500 & 52.22 & 38.33 & 25.28 & 38.61 \\
\bottomrule
\end{tabular}
\end{table}

The source paper reports a GRPO peak within 500 steps of
51.10/38.30/23.70, or 37.70 Macro \citep{zhao2026opsd}. Our local best
single checkpoint is step 300 at 39.44. Selecting the best local checkpoint
separately for each benchmark would produce 56.11/38.61/25.56, or 40.09
Macro, but this is a non-deployable diagnostic oracle and is not used as a
method result. GRPO and OPSD use different objectives and 500- versus
100-step schedules; the curve is therefore kept separate from the matched
Pure--Random--Structured comparison.

\section{Component and Rank Ablations}
\label{app:component-rank}

\begin{table}[H]
\centering
\small
\setlength{\tabcolsep}{4pt}
\caption{Qwen3-1.7B component and rank ablations at step 75 under the common
Think Avg@12 protocol.}
\label{tab:component-rank-summary}
\begin{tabular}{lcrrrr}
\toprule
Variant & Rank & AIME24 & AIME25 & HMMT25 & Macro \\
\midrule
Pure OPSD & -- & 54.17 & 43.89 & 26.39 & 41.48 \\
Matched Random & 64 & 57.78 & 40.70 & 27.36 & 41.94 \\
GapOnly & 64 & 56.39 & 41.11 & 29.72 & 42.41 \\
FullHidden & $d=2{,}048$ & 56.94 & 40.74 & 27.78 & 41.82 \\
\midrule
Structured & 32 & 56.53 & 42.13 & 29.17 & 42.61 \\
Structured & 64 & 58.33 & 42.50 & 29.17 & \textbf{43.33} \\
Structured & 128 & 57.22 & 40.70 & 28.89 & 42.27 \\
\bottomrule
\end{tabular}
\end{table}

FullHidden is +0.34 points above Pure but -0.12 below Random; Structured is
+1.51 above FullHidden. GapOnly is +0.47 above Random and -0.92 below
Structured. Finally, rank 64 is +0.72 and +1.06 above ranks 32 and 128,
respectively. The ordering supports the structured selective-subspace design
within the tested setting.

\section{Offline Factor Diagnostics}
\label{app:gate}

The Qwen3-1.7B audit uses 128 prompts and 4,096 sampled response positions. It
partitions prompts into two complementary folds, fits a factor on one fold, and
evaluates it only on the other. For held-out positions $\mathcal{H}$, define the normalized
privileged-gap score
\begin{equation}
\label{eq:gap-score}
    S_{\mathrm{gap}}(F;\mathcal{H})
    =\frac{1}{|\mathcal{H}|}
      \sum_{t\in\mathcal{H}}
      \frac{\lVert F^\top e_t\rVert_2^2}
           {\operatorname{tr}(F^\top F)}.
\end{equation}
This is the cross-fitted quantity behind the reported 4.40$\times$ ratio.

To measure overlap with the existing output objective, let
$g_t=\nabla_{h_t^S}D_{\mathrm{KL}}(q_t\Vert p_t)$. The output-gradient score is
\begin{equation}
    S_{\mathrm{grad}}(F)
    =\mathbb{E}_t
      \left[
      \frac{\lVert F^\top g_t\rVert_2^2}
           {\operatorname{tr}(F^\top F)}
      \right].
\end{equation}
Finally, with intervention strength $\eta=0.5$, we move the student state along
the projected residual,
\begin{equation}
    h_t^{S,F}=h_t^S-\eta FF^\top e_t,
\end{equation}
and define the exact local KL gain
\begin{equation}
\label{eq:kl-gain}
    \Delta_{\mathrm{KL}}(F)
    =\mathbb{E}_t\left[
      D_{\mathrm{KL}}(q_t\Vert p(\cdot\mid h_t^S))
      -D_{\mathrm{KL}}(q_t\Vert p(\cdot\mid h_t^{S,F}))
    \right].
\end{equation}
The positive-intervention ratio is the fraction of positions whose bracketed
difference in \cref{eq:kl-gain} is positive.

\begin{table}[H]
\centering
\small
\caption{Frozen Qwen3-1.7B offline diagnostics.}
\label{tab:gate-diagnostics}
\begin{tabular}{lrrr}
\toprule
Diagnostic & Structured & Random & Ratio / delta \\
\midrule
A fit $\rightarrow$ B gap & 344.36 & 79.35 & $4.34\times$ \\
B fit $\rightarrow$ A gap & 335.30 & 75.25 & $4.46\times$ \\
Full-data gap / trace & 9.02 & 1.37 & $6.59\times$ \\
Output-gradient / trace & $3.86{\times}10^{-5}$ & $7.43{\times}10^{-5}$ & -48.1\% \\
Exact KL gain & 0.2347 & 0.0241 & $9.75\times$ \\
Positive intervention & 60.55\% & 60.30\% & +0.25 pp \\
\bottomrule
\end{tabular}
\end{table}

The two cross-fitted rank-64 bases have canonical cosine 0.369, indicating
stable held-out score without identical coordinates. Structured intervention
gain is 0.6544, 0.0023, and 0.0264 in early, middle, and late response buckets.
The 4B and 8B pipelines repeat the same integrity and cross-fit checks before
training; none of these diagnostics is used as a substitute for downstream
evaluation.

\subsubsection*{Reproducibility statement}
\Cref{sec:experiments} specifies the training, checkpoint, and evaluation
protocols. Appendix~\ref{app:reproducibility} reports hyperparameters, hardware,
statistical estimands, analytical incremental costs, complete checkpoint curves,
and the offline calibration audit. The accompanying arXiv source bundle contains
the manuscript sources, bibliography, and final figures.

\end{document}